\pdfoutput=1
\documentclass[10pt,twocolumn,letterpaper]{article}
\usepackage[pagenumbers]{cvpr}
\usepackage{booktabs}
\usepackage{multirow}
\usepackage{tabularx}
\usepackage{array}
\usepackage{xcolor}
\usepackage{xurl}
\usepackage{enumitem}
\usepackage{algorithm}
\usepackage{algpseudocode}
\usepackage{fvextra}
\usepackage{placeins}
\usepackage{microtype}
\usepackage{amssymb}
\usepackage{pgfplots}
\pgfplotsset{compat=1.17}
\usetikzlibrary{calc}
\definecolor{c4B}{HTML}{4C72B0}
\definecolor{c8B}{HTML}{DD8452}
\definecolor{cFl}{HTML}{55A868}
\definecolor{c32B}{HTML}{8172B3}
\makeatletter
\newcommand{\teasercaptype}{\def\@captype{figure}}
\makeatother

\newcommand{\pp}{\,pp}
\newcommand{\myparagraph}[1]{{\vspace{.3em} \noindent \bf #1}}

\ifdefined\pdfsuppressptexinfo\pdfsuppressptexinfo=-1\fi

\definecolor{cvprblue}{rgb}{0.21,0.49,0.74}
\usepackage[pagebackref,breaklinks,colorlinks,allcolors=cvprblue]{hyperref}
\def\confName{arXiv}
\def\confYear{2026}
\title{Keep the Image in Mind: A Visual Board for Multimodal Agents}
\author{Haiming Zhu \qquad Zhiwei Chen \qquad Ding Cao \qquad Tianxiang Zhao\\
The Hong Kong University of Science and Technology (Guangzhou)}

\begin{document}
\maketitle
\begin{abstract}
Multi-step visual reasoning suffers when early visual misinterpretations persist as incorrect premises in subsequent reasoning.
We argue that how visual evidence is represented during reasoning matters as much as how much of it is gathered.
Rather than appending observations and inferences to an ever-growing dialogue, we introduce a Visual Board-centered harness that separates visual observations, uncertain interpretations, and derived reasoning within a structured, capacity-bounded, and revisable workspace.
This separation makes each premise explicit and auditable, so that conflicts between what is seen and what is derived become detectable even by a text-only critic and can trigger targeted reinspection of the original image.
Without task-specific training, our harness consistently outperforms standalone models across 4B, 8B, Qwen3-VL-Flash, and 32B configurations on Euclid30K, SolidGeo, and ChartQA bar charts, with gains of up to 11.68, 18.42, and 24.82 percentage points, respectively.
Under identical tools, critic, and interaction limits, the structured Board outperforms shared free-form notes by 6.67 to 13.00 points and no-Board retries by 7.35 to 18.80 points while using 7.9\% to 38.2\% fewer tokens.
Supervised fine-tuning on harness trajectories further improves direct answering and critic-free tool use, with positive transfer to chart reasoning.
\end{abstract}

\begin{figure}[t]
  \centering
  \includegraphics[width=\linewidth]{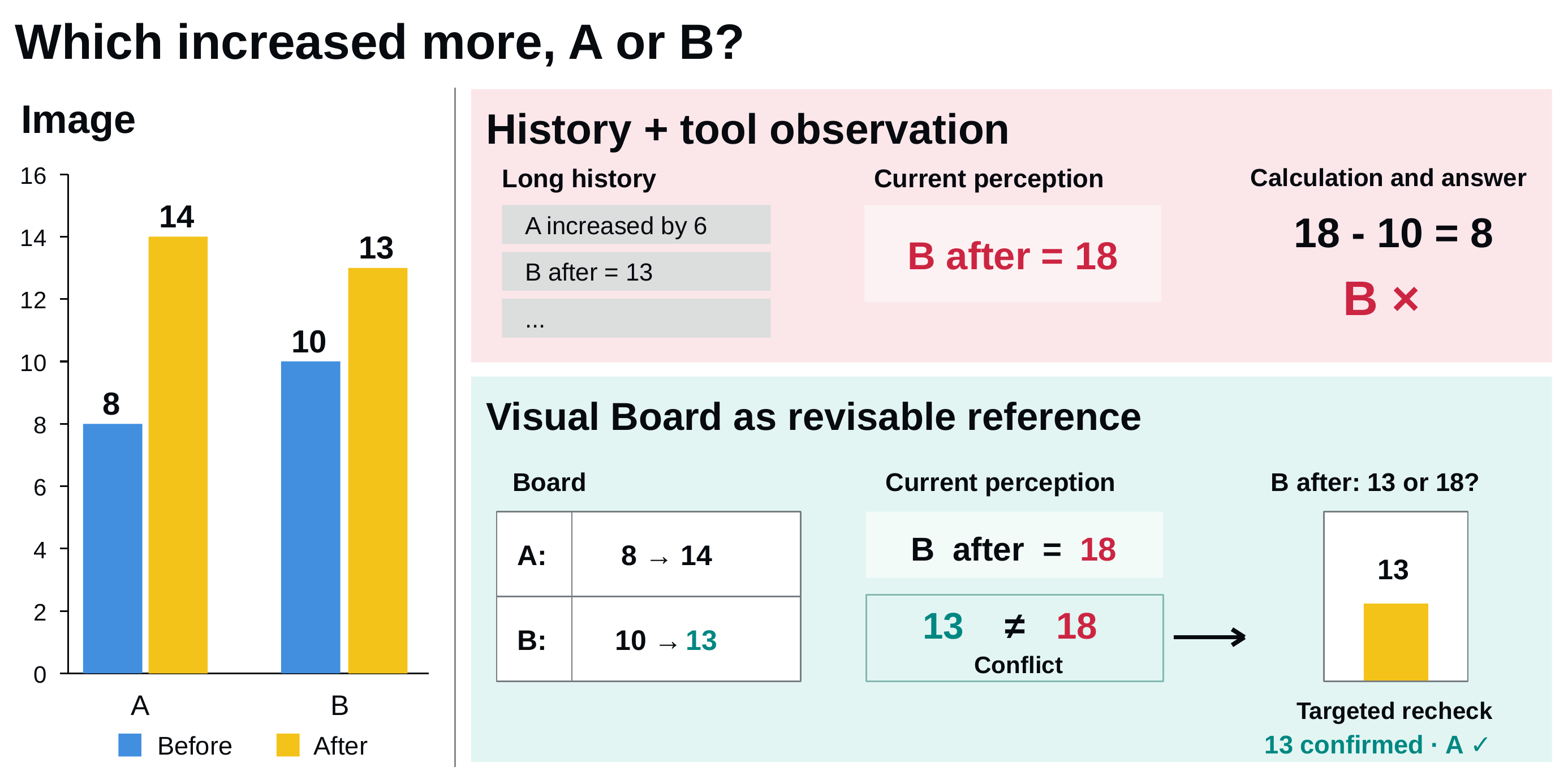}
  \caption{\textbf{The Visual Board as a revisable reference.}
  \textbf{Upper path:} an append-only interaction history carries a mistaken observation forward into every later step.
  \textbf{Lower path:} the Board keeps image facts, uncertain facts, and reasoning steps as explicit, separate premises, so the conflict becomes visible to a text-only Critic and guides targeted reinspection of the original image.}
  \label{fig:visual-board-toy}
\end{figure}

\section{Introduction}
\label{sec:intro}
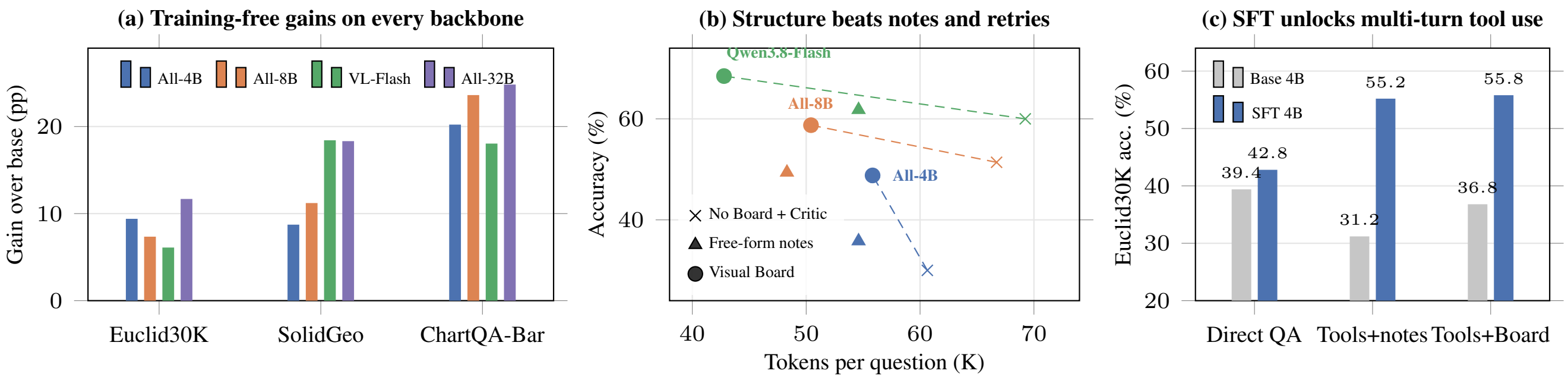
\begin{figure*}[t]
\centering
\resizebox{\textwidth}{!}{%
\begin{tikzpicture}[font=\scriptsize]
\begin{axis}[
  name=pa, width=6.6cm, height=4.3cm,
  ybar, bar width=3.6pt, ymin=0, ymax=29,
  symbolic x coords={Euclid30K,SolidGeo,ChartQA-Bar}, xtick=data,
  enlarge x limits=0.22,
  ylabel={Gain over base (pp)}, ylabel style={yshift=-4pt},
  title={(a) Training-free gains on every backbone},
  title style={font=\scriptsize\bfseries, yshift=-4pt},
  legend style={at={(0.5,0.98)}, anchor=north, legend columns=4,
    draw=none, fill=none, font=\tiny, /tikz/every even column/.append style={column sep=2pt}},
  ymajorgrids, grid style={gray!20}, tick align=outside,
]
\addplot[fill=c4B, draw=none] coordinates {(Euclid30K,9.40) (SolidGeo,8.73) (ChartQA-Bar,20.22)};
\addplot[fill=c8B, draw=none] coordinates {(Euclid30K,7.35) (SolidGeo,11.21) (ChartQA-Bar,23.61)};
\addplot[fill=cFl, draw=none] coordinates {(Euclid30K,6.10) (SolidGeo,18.42) (ChartQA-Bar,18.04)};
\addplot[fill=c32B, draw=none] coordinates {(Euclid30K,11.68) (SolidGeo,18.32) (ChartQA-Bar,24.82)};
\legend{All-4B, All-8B, VL-Flash, All-32B}
\end{axis}
\begin{axis}[
  name=pb, at={($(pa.east)+(1.25cm,0)$)}, anchor=west,
  width=6.0cm, height=4.3cm,
  xmin=38, xmax=74, ymin=24, ymax=74,
  xlabel={Tokens per question (K)}, ylabel={Accuracy (\%)},
  xlabel style={yshift=3pt}, ylabel style={yshift=-4pt},
  title={(b) Structure beats notes and retries},
  title style={font=\scriptsize\bfseries, yshift=-4pt},
  legend style={at={(0.02,0.02)}, anchor=south west, draw=none,
    fill=white, fill opacity=0.8, text opacity=1, font=\tiny},
  legend cell align=left,
  ymajorgrids, xmajorgrids, grid style={gray!20}, tick align=outside,
]
\addlegendimage{only marks, mark=x, mark size=2.4pt, black}\addlegendentry{No Board + Critic}
\addlegendimage{only marks, mark=triangle*, mark size=2.2pt, black}\addlegendentry{Free-form notes}
\addlegendimage{only marks, mark=*, mark size=2.2pt, black}\addlegendentry{Visual Board}
\addplot[c4B, ->, densely dashed, forget plot] coordinates {(60.64,30.00) (55.83,48.80)};
\addplot[c4B, only marks, mark=x, mark size=2.4pt, forget plot] coordinates {(60.64,30.00)};
\addplot[c4B, only marks, mark=triangle*, mark size=2.2pt, forget plot] coordinates {(54.60,35.80)};
\addplot[c4B, only marks, mark=*, mark size=2.2pt, forget plot] coordinates {(55.83,48.80)};
\node[c4B, anchor=west, font=\tiny\bfseries] at (axis cs:56.6,48.8) {All-4B};
\addplot[c8B, ->, densely dashed, forget plot] coordinates {(66.72,51.40) (50.42,58.75)};
\addplot[c8B, only marks, mark=x, mark size=2.4pt, forget plot] coordinates {(66.72,51.40)};
\addplot[c8B, only marks, mark=triangle*, mark size=2.2pt, forget plot] coordinates {(48.31,49.40)};
\addplot[c8B, only marks, mark=*, mark size=2.2pt, forget plot] coordinates {(50.42,58.75)};
\node[c8B, anchor=south, font=\tiny\bfseries] at (axis cs:50.4,59.6) {All-8B};
\addplot[cFl, ->, densely dashed, forget plot] coordinates {(69.24,60.00) (42.78,68.47)};
\addplot[cFl, only marks, mark=x, mark size=2.4pt, forget plot] coordinates {(69.24,60.00)};
\addplot[cFl, only marks, mark=triangle*, mark size=2.2pt, forget plot] coordinates {(54.60,61.80)};
\addplot[cFl, only marks, mark=*, mark size=2.2pt, forget plot] coordinates {(42.78,68.47)};
\node[cFl, anchor=south west, font=\tiny\bfseries] at (axis cs:42.0,69.2) {Qwen3.8-Flash};
\end{axis}
\begin{axis}[
  name=pc, at={($(pb.east)+(1.25cm,0)$)}, anchor=west,
  width=5.4cm, height=4.3cm,
  ybar, bar width=6pt, ymin=20, ymax=64,
  symbolic x coords={Direct QA,Tools+notes,Tools+Board}, xtick=data,
  enlarge x limits=0.25,
  ylabel={Euclid30K acc. (\%)}, ylabel style={yshift=-4pt},
  title={(c) SFT unlocks multi-turn tool use},
  title style={font=\scriptsize\bfseries, yshift=-4pt},
  legend style={at={(0.02,0.98)}, anchor=north west, draw=none, fill=none, font=\tiny},
  nodes near coords, nodes near coords style={font=\tiny, /pgf/number format/fixed, /pgf/number format/precision=1},
  ymajorgrids, grid style={gray!20}, tick align=outside,
]
\addplot[fill=gray!45, draw=none] coordinates {(Direct QA,39.40) (Tools+notes,31.20) (Tools+Board,36.80)};
\addplot[fill=c4B, draw=none] coordinates {(Direct QA,42.80) (Tools+notes,55.20) (Tools+Board,55.80)};
\legend{Base 4B, SFT 4B}
\end{axis}
\end{tikzpicture}}
\caption{\textbf{Overview of results.}
(a)~Without task-specific training, the Visual Board harness improves all four backbone configurations on all three benchmarks over their standalone models.
(b)~On 500 Euclid30K questions with identical tools, Critic, and interaction limits, the structured Board ($\bullet$) is more accurate than shared free-form notes ($\blacktriangle$) and no-Board retries ($\times$) for every backbone, while using fewer tokens than no-Board retries (dashed arrows).
(c)~Supervised fine-tuning of a Qwen3-VL-4B Solver on geometry harness trajectories helps critic-free tool use far more than direct answering.}
\label{fig:results-overview}
\end{figure*}

Answering questions about geometric diagrams~\citep{Lian_2026_CVPR},
spatial configurations~\citep{wang2026solidgeo},
and charts~\citep{masry-etal-2022-chartqa,masry-etal-2025-chartqapro} requires both visual observation and reasoning over what is observed~\citep{su2025thinking}.
A model may need to read a small label, identify a relation,
and combine several observations to derive an answer.
Even a logically coherent derivation can yield an incorrect answer
when it relies on a mistaken visual premise~\citep{Zhang_2026_CVPR}.
Reliable visual reasoning therefore requires not only drawing
conclusions from observations, but also keeping those observations
open to re-examination as reasoning proceeds.

Across interaction steps, visual observations and derived conclusions
are carried forward as generated text~\citep{you2023idealgpt,zhou2025proreason}.
This text provides context for further reasoning, but does not
independently establish that the underlying visual interpretation
is correct.
Studies of multimodal dialogue~\citep{zhong2024investigating,jiang2026mmsnowball} show that misleading textual histories
can cause models to propagate incorrect premises despite access
to the image.
Image-swap experiments~\citep{shi2026are} further show that
self-reflective statements during continuous generation do not
reliably trigger visual re-examination, whereas models are capable
of re-establishing visual grounding when explicitly instructed
in a new dialogue turn.
These findings motivate a clear distinction between revising a
textual reasoning trace and re-examining its underlying visual premises.

A natural response is to make visual inspection an explicit part
of the reasoning process.
Multi-step frameworks~\citep{Xu_2025_ICCV} and visual tools~\citep{zhang2026thyme} allow agents to acquire
missing observations and focus on relevant image
details~\citep{zhou2025proreason,hu2024visual,wu2026vtool,fu2025refocus}.
However, renewed visual access alone does not ensure
that earlier interpretations are reconsidered.
When a visual expert makes an error, downstream agents may
continue to rely on its output rather than question it,
even when a referee points to a possible
mistake~\citep{zhou2025proreason}.
Moreover, additional inspection does not necessarily yield a
reliable basis for revision, as attempts to gather further
information may introduce hallucinations rather than improve
the answer~\citep{kumar2024mmctagent}.
Across this progression, harnesses gather more visual evidence but keep it
in an append-only history, where earlier interpretations are rarely revisited.
The challenge is therefore not only to enable further inspection, but also
to make claims requiring verification explicit and use new evidence to
revise those premises and the reasoning that depends on them.

In contrast, we show that changing how visual evidence is represented, rather than how much of it is gathered, makes multi-step visual reasoning auditable and revisable.
Concretely, we introduce a Visual Board-centered harness for multi-step visual question answering, combining a VLM solver equipped with visual inspection tools and a text-only LLM critic.
Throughout reasoning, the solver maintains a Visual Board that records visual observations, unresolved uncertainties, and derived reasoning in separate fields, while retaining access to the original image.
As illustrated in Fig.~\ref{fig:visual-board-toy}, early visual misinterpretations may persist in the interaction history and influence subsequent reasoning, even when conflicting evidence becomes available.
On the Board, the same premises are explicit, so a conflict between an observation and a derivation becomes visible and can direct targeted reinspection of the original image.

Formally, a standard multi-step harness appends each new observation or inference to the interaction history, $h_{t+1}=h_t\oplus u_t$, so an early misreading remains a premise of every later step.
In contrast, the Board makes the reasoning state an explicit object that is rewritten rather than extended, $B_{t+1}=g(B_t,u_t)$ with $B_t=(\mathcal{F}_t,\mathcal{U}_t,\mathcal{R}_t)$, where $\mathcal{F}_t$ holds image facts, $\mathcal{U}_t$ uncertain image facts, and $\mathcal{R}_t$ reasoning steps, and $u_t$ collects tool outputs and Critic feedback.
Because the Board is capacity-bounded and updated by snapshot replacement, it does not grow with the dialogue.
Separating what is seen from what is derived keeps unverified interpretations from being treated as visual facts, lets a text-only critic check whether each conclusion follows from the recorded premises, and reduces interference from accumulated history, with the original image as the authoritative source.

We evaluate our harness on Euclid30K, SolidGeo, and ChartQA, spanning geometric, spatial, and chart reasoning.
Without task-specific training, the harness improves all four backbones on all three benchmarks by 6.10 to 24.82 points (Fig.~\ref{fig:results-overview}a), and the homogeneous All-8B system (58.75\%, 47.20\%, 81.60\%) outperforms standalone Qwen3-VL-32B on every benchmark with a four-times smaller backbone.
Controlled ablations on 500 Euclid30K questions isolate the representation itself by fixing the solver, tools, critic, and interaction limits.
The structured Board outperforms shared free-form notes by 13.00, 9.35, and 6.67 points on All-4B, All-8B, and Qwen3.8-Flash, with a paired 95\% bootstrap interval of $[8.43, 17.67]$ points on the 498 common All-4B questions.
It also outperforms no-Board retries by 18.80, 7.35, and 8.47 points while using 7.9\%, 24.4\%, and 38.2\% fewer tokens, so the gains do not come from additional computation.
Supervised fine-tuning on harness trajectories further improves direct answering and tool-augmented inference, with positive transfer from geometry to charts.

\noindent We summarize our main contributions in three parts.
\begin{itemize}
    \item We show that the representation of visual evidence matters:
    the Visual Board, a structured, capacity-bounded, and revisable
    workspace, decouples image observations and visual uncertainties
    from derived reasoning, and outperforms shared free-form notes and
    no-Board retries under identical tools, critic, and limits.
    Instead of growing with the dialogue, it holds past and future
    visual updates in one structured state through snapshot
    replacement, which distinguishes it from memory and history
    compression despite similarities in form.

    \item We design selective board construction,
    board-preserving history compression, and query-driven
    visual inspection to support premise revision under
    compact textual context. These mechanisms preserve access
    to the original image and allow uncertain details and
    conflicting claims to guide targeted re-examination of the image.

    \item We show that the gains are training-free and broad:
    across Euclid30K, SolidGeo, and ChartQA and multiple model scales
    the harness improves every backbone, and a text-only critic helps
    most for smaller solvers.
    SFT on harness trajectories further improves both model-only and
    harness-based inference, with transfer from geometry to charts.
\end{itemize}

\section{Related Work}
\label{sec:related}

\myparagraph{Multi-agent visual reasoning.}
A line of work separates visual perception from language-based
reasoning through iterative model collaboration.
IdealGPT~\citep{you2023idealgpt} decomposes questions into visual
sub-questions and combines their answers through iterative reasoning.
ProReason~\citep{zhou2025proreason} decouples visual perception
from textual reasoning, coordinating multiple agents to proactively
gather task-relevant information and assess its sufficiency
before answering.
SeeingEye~\citep{zhang2025seeingeyeagenticinformationflow} uses a
tool-equipped VLM to construct question-conditioned structured
representations for a text-only LLM, while
BeMyEyes~\citep{huang2025eyesextendinglargelanguage} trains a perceiver
to collaborate with a reasoner through multi-turn dialogue.
DRP~\citep{jia2026mitigating} addresses visual context degradation
by allowing a text-only reasoner to request visual observations
on demand.
These methods decouple perception and reasoning across \emph{models};
we decouple them within the \emph{representation}, keeping observations,
uncertainties, and derived reasoning in separate, replaceable Board
fields for an image-conditioned Solver.

\myparagraph{Context management for visual reasoning.}
Several approaches explicitly organize intermediate visual
information to support subsequent reasoning.
MMCTAgent~\citep{kumar2024mmctagent} introduces a vision-based
critic to evaluate reasoning and visual evidence.
MSSR~\citep{guo2026pursuing} constructs a compact, sufficient set
of 3D perception results for spatial question answering.
See It, Say It, Sorted~\citep{Zhang_2026_CVPR} maintains a growing
textual visual-evidence pool to guide token-level decoding and
requests additional observations when existing evidence is
insufficient.
Complementing these approaches, CoSee~\citep{zhou2026diagnosing}
examines how shared notes can reinforce visual errors in
resource-constrained agents and evaluates verification before
state commitment on ChartQAPro.
Agent memory~\citep{packer2023memgpt,park2023generative,fan2024videoagent}
and prompt compression~\citep{jiang2023llmlingua} instead retain,
retrieve, or condense past interaction to extend usable context.
Recent text-agent work moves from harness-scheduled compaction toward
model-controlled context, letting the model edit its own live context
as a file~\citep{shao2026clm}.
These methods manage context for long-horizon text agents with a
general-purpose, unconstrained edit space; we instead impose a
visual-specific schema (observations, uncertainties, derivations)
so that a text-only critic can audit premises, and keep the original
image as the authoritative evidence.
Our Visual Board neither stores nor summarizes history; it is a
capacity-bounded workspace revised by snapshot replacement, with
uncertainties recorded to guide later queries.

\myparagraph{Tool-augmented visual reasoning.}
Tool-augmented approaches extend reasoning with explicit visual
operations and external feedback.
Visual Sketchpad~\citep{hu2024visual} supports reasoning through
drawing, while VisuoThink~\citep{wang2025visuothink} interleaves
visual and textual reasoning with look-ahead tree search.
Draft and Refine~\citep{jeong2026draft} selects visual experts
using a question-conditioned utilization metric, and
ForeSight~\citep{Wu_2026_CVPR} combines low-level visual cues
with mask-based feedback to support answer revision.
Our harness complements these approaches by using the Visual
Board to track visual uncertainties across reasoning steps
and guide targeted re-inspection.
Both crop-based inspection and whole-image queries require
formulating an explicit, neutral visual question.
Unlike pipelines that append tool outputs to the dialogue, we treat
them as uncommitted proposals that the Solver accepts into the Board.

\myparagraph{Post-training for visual reasoning.}
Recent work trains VLMs to coordinate reasoning and visual actions~\citep{shi2025mathcanvas}.
AdaReasoner~\citep{song2026adareasoner} introduces Tool-GRPO,
extending GRPO~\citep{deepseek-math} to optimize multi-turn
tool-calling trajectories, while
VTool-R1~\citep{wu2026vtool} trains VLMs to interleave textual
reasoning with intermediate images produced by Python-based
visual editing.
VISTA-Gym~\citep{lu2026scaling} provides a standardized interactive
environment for training tool-integrated visual reasoning.
Learning also improves perception-reasoning interfaces, as
RAPID~\citep{gou2026reasoning} optimizes visual descriptions using
an external reasoner's answer correctness, and
VALOR~\citep{marsili2026no} combines LLM-based verification for
reasoning reinforcement learning with VLM-based hard-negative
mining for visual grounding.
Through region-to-image distillation, ZwZ~\citep{wei2026zooming} improves fine-grained perception without test-time tool use.
We further investigate whether supervised fine-tuning enables
a 4B VLM solver to use the Visual Board more effectively
in multi-step visual reasoning.

\section{Method}
\label{sec:method}
\begin{figure*}[t]
  \centering
  \includegraphics[width=\textwidth]{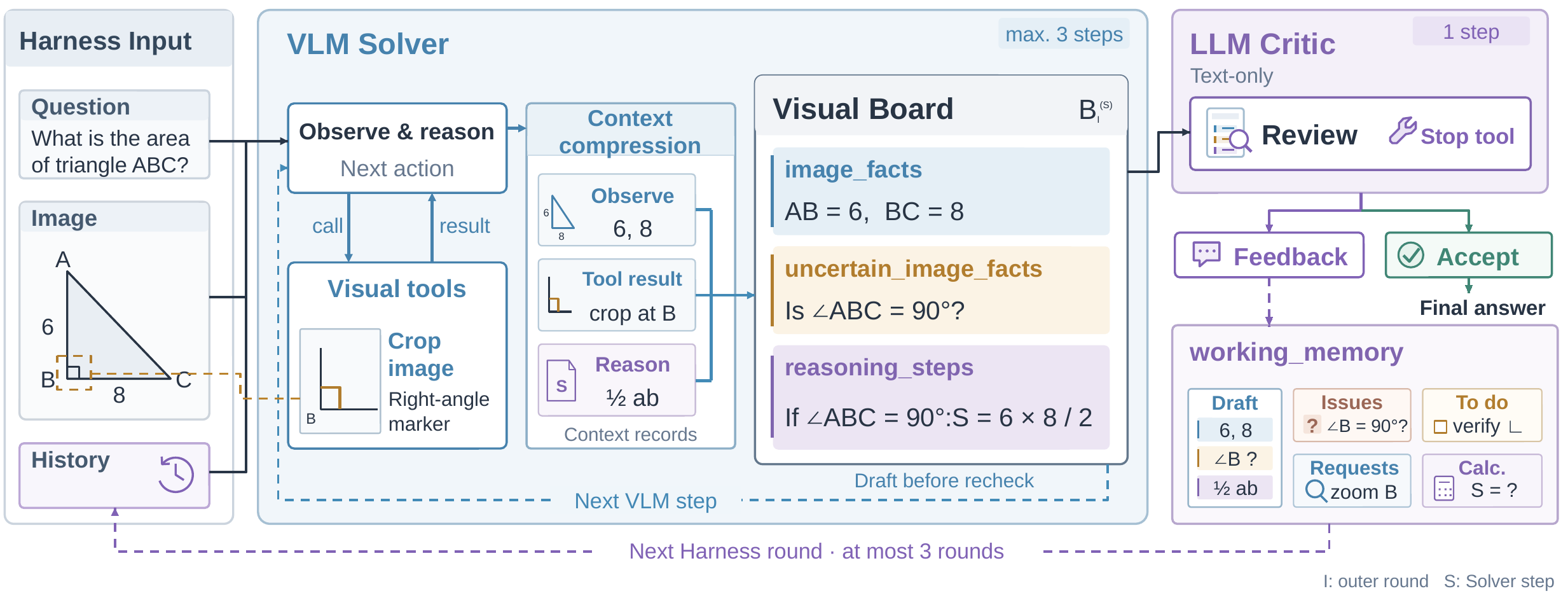}
  \caption{\textbf{Overview of the Visual Board-centered harness.}
  The VLM Solver keeps the original image available, maintains a revisable
  Board, and invokes focused crop or whole-image queries whenever a visual premise
  requires checking. A text-only Critic reviews the submitted premises and
  derivation to return structured feedback, without directly inspecting the image.
  History compression preserves both the current Board and direct image access.}
  \label{fig:pipeline}
\end{figure*}

\myparagraph{Overall pipeline.}
Our harness follows a multi-round solver-critic workflow
illustrated in Fig.~\ref{fig:pipeline}. At round $I$, an image-conditioned VLM Solver
receives the image $\mathcal{I}$, question $Q$, current Visual Board,
and feedback from the previous round, when available.
It reasons over the image, optionally invokes visual or computation
tools, and submits an answer or a request for missing visual evidence.
A text-only LLM Critic checks the submitted reasoning and returns
feedback for a further round. The protocol runs for up to three outer
rounds with a fixed per-round Solver-step limit, and the supplement
gives the full algorithm and prompts.

\subsection{The Visual Board}
\label{subsec:board}
The Board is a structured linguistic representation of the visual
premises currently used in reasoning, rather than an append-only
transcript or dialogue memory. It decouples what is seen from what is
derived and does not incrementally extend image content.
At Solver step $S$ of round $I$, the Board state is given by
\begin{equation}
 B_I^S = \bigl(F_I^{S,\mathrm{img}},U_I^{S,\mathrm{img}},R_I^S\bigr).
\end{equation}
The three fields serve distinct roles within each round.
\begin{itemize}[leftmargin=*,nosep]
\item $F^{\mathrm{img}}$, denoted \texttt{image\_facts}, records the current
interpretation of answer-relevant image content.
\item $U^{\mathrm{img}}$, denoted \texttt{uncertain\_image\_facts}, records
unresolved visible details requiring further inspection.
\item $R$, denoted \texttt{reasoning\_steps}, records derived relations,
calculations, and intermediate conclusions.
\end{itemize}
Question-given constraints are maintained separately by the harness.
In particular, a derived relation is not promoted to an observed image
fact merely because it proves useful while solving the problem.

The Solver keeps compact step-level working notes, labeled \texttt{working\_memory}
in Fig.~\ref{fig:pipeline}, during intermediate steps
and submits complete snapshots of the three fields through terminal
tool calls. A subsequent snapshot can replace a mistaken observation,
withdraw a claim into uncertainty, or update reasoning that depends on
the changed premise. New tool observations are proposals for revision,
not automatic edits to the committed Board, and each carries a bounded
number of candidate entries. Because snapshots replace rather than
extend the Board, its size tracks the current task state rather than
the interaction length, without retaining superseded assertions.

\myparagraph{Evidence remains provisional.}
Entries in \texttt{image\_facts} represent provisional visual hypotheses
rather than fixed ground truth. All Board entries and tool observations
remain revisable, while the original image serves as the primary visual
evidence. Unverified relations are marked as assumptions so conclusions
depending on them remain conditional. Beyond error correction, the Board
also organizes evidence throughout multi-step solving.

\subsection{Visual Inspection and Reasoning Feedback}
\label{subsec:tools}
\myparagraph{Query-driven tools.}
Both visual tools receive a neutral, observation-oriented query,
together with the original question and any answer options.
\texttt{grid\_caption} inspects a Solver-specified crop
$b=(x_1,y_1,x_2,y_2)\in[0,1]^4$, for example to read a small label
or check a diagram marker. \texttt{vlm\_query} inspects the full image
when the relevant location is uncertain or a relation spans regions.
Python execution supports calculations from already available premises.
Visual queries must not embed an expected answer or ask the tool to
perform a derivation. A repeated inspection must change the region,
tool, or requested fact when the earlier observation is insufficient
or contradictory.

\myparagraph{Structured submission.}
The terminal tools transfer the current solution state to the Critic.
\texttt{terminate\_and\_answer} submits an answer, confidence,
complete Board snapshots, and zero-based indices of supporting
\texttt{image\_facts} entries.
\texttt{need\_more\_visual\_observation} instead specifies the missing
visual detail, its location and relevance, and the visual tool required next.
Before accepting a subsequent terminal submission after a visual
handoff, the harness requires execution of the designated
tool with a structured observation result. Tool responses acknowledge
submissions without duplicating the committed Board state.

\myparagraph{Reasoning critique.}
The Critic receives the question, submitted Board, answer, and textual
trace. It checks whether the derivation follows from the recorded
premises and flags concrete, answer-critical inconsistencies or
unsupported assumptions. Without access to the image, it neither
solves the task independently nor edits the Board; instead, its feedback
can request a targeted visual recheck while leaving visual verification
and revision to the VLM Solver. As examined in Sec.~\ref{subsec:critic},
the empirical benefit of Critic feedback varies with backbone strength.
In Flash configurations, an optional post-submission visual-backend
gate checks cited \texttt{image\_facts} entries against the image
before an answer is accepted, as detailed in the supplement.

\subsection{Board-Preserving History Compression}
\label{subsec:compression}
The harness limits accumulated context using deterministic rules,
without additional model calls or learned semantic summarization.
It bounds individual observation records on entry, retires completed
step prompts, and selectively reconstructs older interaction history.
Compression acts only on transient history; the Board is never a
summary of it and is carried separately, while the original image
remains directly accessible to the Solver in every round.

Under default thresholds, reconstruction triggers when
\begin{equation}
 N_{\mathrm{msg}}(H_I^S)>7
 \quad\lor\quad N_{\mathrm{char}}(H_I^S)>4500,
\end{equation}
where the character count excludes image data and tool-call arguments.
When triggered, the reconstructed history retains the original image-bearing message
and the latest tool call with its responses. Older tool outputs are
replaced by compact, type-specific records; earlier free-form reasoning
and previous summaries need not be retained. Structured visual updates
remain marked as uncommitted proposals until the Solver incorporates
them. Detailed selection rules and length limits are in the supplement.

\subsection{Harness-Aware Supervised Fine-Tuning}
\label{subsec:4btraing}

We initialize the student Solver from Qwen3-VL-4B-Instruct~\citep{bai2025qwen3vltechnicalreport} and train it on decision-level examples extracted from 1,556 correct, protocol-valid trajectories collected from the Euclid30K training split.
These trajectories are generated using Qwen3-VL-Flash as the Solver and visual-tool backend, with Qwen3-Max serving as the Critic.
Each training example pairs a teacher response with its decision-time context, including the image, question, current Visual Board, retained observations, feedback, and available tool schemas.
To prevent data leakage, source questions are partitioned before trajectory expansion, ensuring that decisions from the same question remain within a single split.

For input $x=(\mathcal{I},Q,\widetilde H,\mathcal T)$ and target Solver
response $y$, including any serialized tool calls, we minimize
\begin{equation}
 \mathcal L_{\mathrm{SFT}}(\theta)=
 -\sum_{(x,y)\in\mathcal D_{\mathrm{SFT}}}
 \sum_{j=1}^{|y|}\log\pi_\theta(y_j\mid x,y_{<j}).
\end{equation}
where $\widetilde H$ is the retained, compressed interaction history and
$\mathcal T$ is the set of available tool schemas.
Only the current response contributes to the loss. Historical replies,
tool outputs, and Critic feedback are conditioning context with masked
labels. This trains multi-step interaction alongside answering, and
our experiments separate training gains from imposing a specific Board format.

\begin{table*}[t]
\centering

\caption{
\textbf{Training-free performance compared with standalone backbones.}
We report accuracy in percent and improvements $\Delta$ in percentage points on Euclid30K, SolidGeo, and ChartQA-Bar~\citep{wu2026vtool}, comprising 500, 2,206, and 826 questions, respectively.
All-4B, All-8B, and All-32B use size-matched models for the Solver, visual tools, and Critic, whereas Qwen3-VL-Flash employs Qwen3-Max as its Critic.
Boldface indicates the higher accuracy in each comparison.
}

\label{tab:comparison}
\small
\setlength{\tabcolsep}{3pt}
\begin{tabular*}{\textwidth}{@{\extracolsep{\fill}}lrrrrrrrrr@{}}
\toprule
 & \multicolumn{3}{c}{Euclid30K} & \multicolumn{3}{c}{SolidGeo} & \multicolumn{3}{c}{ChartQA-Bar} \\
\cmidrule(lr){2-4}\cmidrule(lr){5-7}\cmidrule(l){8-10}
System / Backbone & Base & Board & $\Delta$ & Base & Board & $\Delta$ & Base & Board & $\Delta$ \\
\midrule
All-4B / Qwen3-VL-4B & 39.40 & \textbf{48.80} & +9.40 & 32.11 & \textbf{40.84} & +8.73 & 57.75 & \textbf{77.97} & +20.22 \\
All-8B / Qwen3-VL-8B & 51.40 & \textbf{58.75} & +7.35 & 35.99 & \textbf{47.20} & +11.21 & 57.99 & \textbf{81.60} & +23.61 \\
Qwen3-VL-Flash & 50.80 & \textbf{56.90} & +6.10 & 39.83 & \textbf{58.25} & +18.42 & 68.52 & \textbf{86.56} & +18.04 \\
All-32B / Qwen3-VL-32B & 57.40 & \textbf{69.08} & +11.68 & 41.02 & \textbf{59.34} & +18.32 & 61.86 & \textbf{86.68} & +24.82 \\
\bottomrule
\end{tabular*}
\end{table*}

\section{Experiments}
\label{sec:exp}
We first compare the harness against each backbone's standalone base model,
and then examine Board structure, Critic feedback, and SFT transfer to charts.

\subsection{Experimental Setup}
\label{subsec:setup}

\myparagraph{Benchmarks.}
Euclid30K~\citep{Lian_2026_CVPR} evaluates plane and solid geometry on 500 held-out questions, disjoint from the SFT training split.
SolidGeo~\citep{wang2026solidgeo} covers projection, unfolding, spatial measurement, and spatial vectors, with 2,206 single-image questions used for evaluation.
ChartQA-Bar refers to the 826-question bar-chart subset of ChartQA~\citep{masry-etal-2022-chartqa} used by VTool-R1~\citep{wu2026vtool}, comprising 444 horizontal and 382 vertical bar charts, each paired with one question.

\myparagraph{Models and controls.}
For the All-4B, All-8B, and All-32B systems, Qwen3-VL-Instruct
of the specified size serves both the Solver and visual tools, with
a correspondingly sized Qwen3 text model as Critic; thus All-4B uses
Qwen3-VL-4B-Instruct and Qwen3-4B without any larger helper model.
In Tab.~\ref{tab:comparison}, the Qwen3-VL-Flash system pairs
Qwen3-VL-Flash for the Solver and visual tools with Qwen3-Max as Critic.
In the 500-question Euclid30K ablations, Qwen3.8-Flash denotes a
homogeneous API configuration using \texttt{qwen3.8-flash} across Solver,
visual tools, and text-only Critic with no image supplied to the Critic.

\myparagraph{Evaluation protocol.}
Model-only baselines follow the official evaluation pipelines of
Euclid30K~\citep{Lian_2026_CVPR}, SolidGeo~\citep{wang2026solidgeo}, and
VTool-R1~\citep{wu2026vtool} with a 4,096-token output cap.
On geometry benchmarks, mathematical answers in equivalent forms such as
\texttt{1/2} and \texttt{0.5} are evaluated via symbolic and
numeric equivalence; on ChartQA-Bar, answers follow standard relaxed matching
with text normalization, unordered list matching, and a 5\% relative numeric tolerance.

\myparagraph{Training setup.}
We evaluate the checkpoint after 200 training steps on the 1,556
geometry trajectories from the Euclid30K training split described in Sec.~\ref{subsec:4btraing}.
Training uses four A800 GPUs, BF16, ZeRO-3, and gradient checkpointing,
with learning rate $10^{-5}$ for the language model and multimodal
aligner and a frozen vision encoder. Over-length decision examples
are discarded rather than truncated. Training uses only geometry
trajectories and has zero overlap with ChartQA-Bar, which serves as our
cross-domain transfer benchmark, with decontamination details in the supplement.

\subsection{Comparison with Base Models}
\begin{table}[t]
\centering
\caption{\textbf{Training-free 4B tool-use control.}
Tools only uses the base 4B Solver and 4B visual backend without
a Board, Critic, or post-submission visual check, whereas the full All-4B
harness retains both the Board and Critic. All scores are accuracies in percent, obtained from base models without any SFT.}
\label{tab:tools-only}
\small
\setlength{\tabcolsep}{3pt}
\begin{tabular*}{\linewidth}{@{\extracolsep{\fill}}lrrr@{}}
\toprule
Setting & Euclid30K & SolidGeo & ChartQA-Bar \\
\midrule
Standalone & 39.40 & 32.11 & 57.75 \\
Tools only & 35.60 & 24.25 & 74.80 \\
Full harness & \textbf{48.80} & \textbf{40.84} & \textbf{77.97} \\
\bottomrule
\end{tabular*}
\end{table}

\begin{table}[t]
\centering
\caption{\textbf{ProReason comparison on Euclid30K.}
Both methods use Qwen3-VL-8B-Instruct and Qwen3-8B on all 500 questions.}
\label{tab:pro-reason-comparison}
\small
\begin{tabular*}{\linewidth}{@{\extracolsep{\fill}}lrr@{}}
\toprule
Method & Acc. (\%) & $\Delta$ (pp) \\
\midrule
ProReason & 51.60 & -- \\
Visual Board & \textbf{58.75} & +7.15 \\
\bottomrule
\end{tabular*}
\end{table}

Tab.~\ref{tab:comparison} compares each harness configuration with its
corresponding model-only baseline. Across all four backbones and three
benchmarks, the training-free harness consistently improves accuracy by
6.10 to 24.82\pp.
On Euclid30K, accuracy increases by 9.40, 7.35, 6.10, and
\textbf{11.68}\pp{} for All-4B, All-8B, Qwen3-VL-Flash, and All-32B,
respectively; on SolidGeo, the corresponding gains are 8.73, 11.21,
\textbf{18.42}, and 18.32\pp.
On ChartQA-Bar, gains reach 20.22, 23.61, 18.04, and \textbf{24.82}\pp{}
across the four model scales.
Gains remain substantial across scales (9.40, 7.35, 6.10, and 11.68\pp{} on Euclid30K), with All-32B obtaining the largest gains on
Euclid30K and ChartQA-Bar, so the harness complements rather than replaces
backbone capacity.

\myparagraph{The harness narrows model-size gaps: All-8B outperforms standalone 32B.}
Structured interaction narrows model-size gaps without requiring larger helper models.
The homogeneous All-4B system achieves 48.80\%, 40.84\%, and 77.97\% on Euclid30K, SolidGeo, and ChartQA-Bar, respectively.
It surpasses standalone Qwen3-VL-8B by 4.85\pp{} on SolidGeo and 19.98\pp{} on ChartQA-Bar, and exceeds standalone Qwen3-VL-32B by 16.11\pp{} on ChartQA-Bar while nearly matching its SolidGeo accuracy of 41.02\%.
Similarly, homogeneous All-8B achieves 58.75\%, 47.20\%, and 81.60\%, outperforming standalone Qwen3-VL-32B, which scores 57.40\%, 41.02\%, and 61.86\%, despite using a four-times smaller backbone.

The largest gains occur on ChartQA-Bar, ranging from 18.04 to 24.82\pp, where crop-guided inspection helps resolve dense bar heights and axis ticks.
SolidGeo also shows substantial improvements of 8.73 to 18.42\pp, suggesting that separating visual observations from spatial derivations helps mitigate errors in 3D projection, unfolding, and vector reasoning across all four backbones.

\myparagraph{Tools alone are not enough for multi-step geometry.}
Tab.~\ref{tab:tools-only} adds a tools-only control using the base
Qwen3-VL-4B for both the Solver and visual backend, with no SFT,
Board, Critic, or post-answer visual checking. It reaches 35.60\%,
24.25\%, and 74.80\% on Euclid30K, SolidGeo, and ChartQA-Bar.
Compared with standalone 4B, raw tool access improves ChartQA-Bar by
+17.05\pp{} where local crop readings map directly to bar values, yet
degrades Euclid30K by $-3.80$\pp{} and SolidGeo by $-7.86$\pp{} because
unorganized multi-turn tool outputs accumulate conflicting geometric claims.
Adding the Board and Critic reverses both drops, outperforming the
tools-only control by +13.20\pp{} on Euclid30K and +16.59\pp{} on SolidGeo
with the same 4B models. Tool access therefore helps when evidence is local,
but multi-step geometry additionally requires a structured reasoning state
that organizes the accumulated observations.

\begin{figure*}[t]
  \centering
  \includegraphics[width=\linewidth]{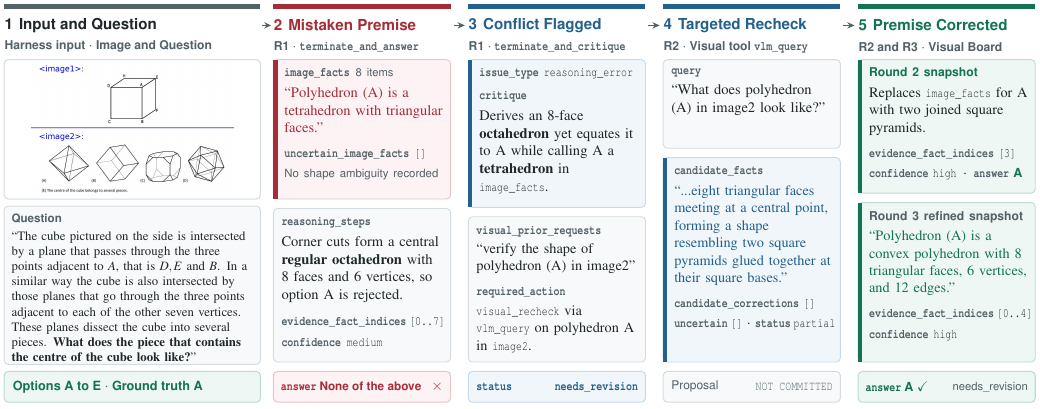}

\caption{
\textbf{Illustrative Euclid30K trajectory under All-8B.}
In Round 1, \texttt{reasoning\_steps} derives a central octahedron while \texttt{image\_facts} records option A as a tetrahedron, giving ``None of the above.''
Because the two occupy separate Board fields, the text-only Critic detects the conflict and requests a \texttt{vlm\_query} recheck.
Round 2 replaces the entry and answers A, and Round 3 refines it.
The final \texttt{needs\_revision} status comes from the post-submission check, and the run ends at the round limit with answer A.
}

  \label{fig:qualitative-recovery}
\end{figure*}

\myparagraph{ProReason: revising premises vs.\ accumulating descriptions.}
We reproduce ProReason~\citep{zhou2025proreason} using
Qwen3-VL-8B-Instruct and Qwen3-8B on held-out Euclid30K.
Tab.~\ref{tab:pro-reason-comparison} reports 51.60\% for ProReason
and 58.75\% for our harness under the same backbone pair,
a 7.15\pp{} improvement.
ProReason separates seeing from reasoning across agents and accumulates
visual descriptions and insights in one memory, so its text-only
summarizer cannot recheck a wrong description against the image.
Our VLM both inspects and answers, and replaces mistaken Board entries
after reinspection, while our Critic never answers the question itself and
only flags unsupported steps for the Solver to recheck against the image.

\subsection{Structured Board versus Free-Form Notes}
\label{sec:ablations}

\begin{table}[t]
\centering
\caption{\textbf{Board ablations on 500-question Euclid30K evaluations.}
Models and interaction limits are fixed within each group. Accuracy counts
failed runs as incorrect, and tokens and time are per-question means.
Bold marks the best accuracy in each group.}
\label{tab:ablation}
\small
\setlength{\tabcolsep}{3pt}
\begin{tabular*}{\linewidth}{@{\extracolsep{\fill}}lrrr@{}}
\toprule
Board setting & Acc. (\%) & Tokens & Time (s) \\
\midrule
\multicolumn{4}{@{}l}{\textit{All-4B}} \\
No Board + Critic & 30.00 & 60,636.13 & 64.73 \\
Shared free-form notes & 35.80 & 54,601.00 & 64.61 \\
Concise Board & 40.40 & 77,257.18 & 83.46 \\
Default Visual Board & \textbf{48.80} & 55,825.19 & 56.87 \\
\midrule
\multicolumn{4}{@{}l}{\textit{All-8B}} \\
No Board + Critic & 51.40 & 66,722.68 & 153.56 \\
Shared free-form notes & 49.40 & 48,313.90 & 112.33 \\
Concise Board & 58.20 & 68,391.39 & 160.86 \\
Default Visual Board & \textbf{58.75} & 50,421.00 & 103.16 \\
\midrule
\multicolumn{4}{@{}l}{\textit{Qwen3.8-Flash}} \\
No Board + Critic & 60.00 & 69,240.22 & 85.56 \\
Shared free-form notes & 61.80 & 54,604.49 & 64.36 \\
Concise Board & 67.20 & 62,733.70 & 68.50 \\
Default Visual Board & \textbf{68.47} & 42,775.75 & 61.66 \\
\bottomrule
\end{tabular*}
\end{table}

\begin{table}[tbp]
\centering
\caption{\textbf{Paired All-4B outcomes on 498 common Euclid30K questions.}
The exact gain is $100(105-40)/498=13.05$\pp, with a 95\% paired
bootstrap interval of $[8.43, 17.67]$\pp.}
\label{tab:paired}
\small
\begin{tabular*}{\linewidth}{@{\extracolsep{\fill}}lrr@{}}
\toprule
 & Free-form correct & Free-form wrong \\
\midrule
Default correct & 139 & 105 \\
Default wrong & 40 & 214 \\
\bottomrule
\end{tabular*}
\end{table}

We study 500-question Euclid30K evaluations across All-4B, All-8B,
and Qwen3.8-Flash in Tab.~\ref{tab:ablation}.
\emph{No Board + Critic} removes the Board but keeps the tools and
Critic-driven retry, unlike the tools-only control above.
\emph{Shared free-form notes} gives the Solver a persistent single-string
text workspace that is shared verbatim with the Critic across rounds.
The model freely writes and updates these notes, choosing their content,
format, and length without separating \texttt{image\_facts},
\texttt{uncertain\_image\_facts}, and \texttt{reasoning\_steps} or citing
indexed premises. This variant isolates structural decoupling from
persistence and sharing alone.
\emph{Concise Board} uses an additional shortened Board prompt.
All Board variants retain Critic feedback, the same helper models,
maximum of three rounds, tool-call limits, and output-token cap,
while prompt and verification details appear in the supplementary material.

\myparagraph{Structure, not persistence, drives the gain.}
The default Board outperforms shared free-form notes by 13.00\pp{}
on All-4B, 9.35\pp{} on All-8B, and 6.67\pp{} on Flash.
On the 498 All-4B questions common to both runs, the Board alone solves 105
and free-form notes alone solve 40 in Tab.~\ref{tab:paired}, a paired
difference of 13.05\pp{} with a 95\% bootstrap interval of $[8.43, 17.67]$\pp{}
over 100,000 resamples.

\begin{figure*}[!t]
  \centering
  \includegraphics[width=\linewidth]{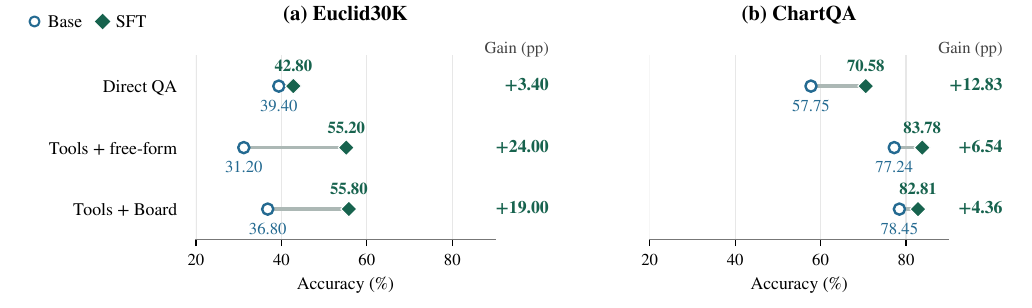}
  \caption{\textbf{Geometry-trained SFT improves Direct QA and critic-free tool use.}
  Within each row, Base and SFT Qwen3-VL-4B Solvers share identical tool access,
  visual-tool backend, and Board format with the Critic disabled to isolate
  Solver fine-tuning. Board denotes the default Visual Board, and ChartQA-Bar
  denotes the 826-item bar-chart split. All SFT evaluations use the same 200-step
  checkpoint; complete numerical values are listed in the supplementary material.}
  \label{fig:sft-transfer}
\end{figure*}

\myparagraph{Terser Boards are neither better nor cheaper.}
The concise variant reaches 40.40\%, 58.20\%, and 67.20\% for
All-4B, All-8B, and Flash, respectively, compared with 48.80\%,
58.75\%, and 68.47\% for the default Board, while consuming more
tokens in every group because overly terse entries omit key premises
and trigger extra review rounds from the Critic.

\myparagraph{The Board beats no-Board retries at lower cost.}
The default Board improves over No Board + Critic by 18.80, 7.35,
and 8.47\pp{} while using 7.9\%, 24.4\%, and 38.2\% fewer
tokens, respectively. Against free-form notes, its token use is
2.2\% and 4.4\% higher for All-4B and All-8B and 21.7\%
lower for Flash, with lower mean elapsed time in all three groups.
Role-level API call breakdowns in the supplement explain why structured
decoupling reduces latency. Without a Board, the Solver and Critic fall into
repeated retry loops over implicit or drifting premises, issuing 4,293 Critic
calls on All-4B and 3,555 on All-8B. The default Board cuts Critic calls by
60.5\% to 1,696 on All-4B and by 55.7\% to 1,575 on All-8B. On All-4B, explicit
\texttt{uncertain\_image\_facts} shifts compute from Critic retries to
targeted visual queries, raising visual calls from 818 to 4,044.
Concise Board triggers over twice as many Critic calls as the default Board,
reaching 3,609 on All-4B and 3,616 on All-8B, because terse snapshots omit
premises needed for single-round review.

\begin{table}[t]
\centering
\caption{\textbf{Critic ablation with a Visual Board retained.}
Each pair fixes the Solver and visual backend on the 500 Euclid30K questions.
Bold marks higher accuracy within each pair.}
\label{tab:critic}
\small
\setlength{\tabcolsep}{3pt}
\begin{tabular*}{\linewidth}{@{\extracolsep{\fill}}llrrr@{}}
\toprule
Configuration & Critic & Acc. (\%) & Tokens & Time (s) \\
\midrule
\multirow{2}{*}{All-4B}
 & With & \textbf{48.80} & 55,825.19 & 56.87 \\
 & Without & 36.80 & 44,274.37 & 43.79 \\
\midrule
\multirow{2}{*}{All-8B}
 & With & \textbf{58.75} & 50,421.00 & 103.16 \\
 & Without & 50.80 & 28,374.05 & 77.36 \\
\midrule
\multirow{2}{*}{Qwen3.8-Flash}
 & With & 68.47 & 42,775.75 & 61.66 \\
 & Without & \textbf{69.80} & 17,187.12 & 26.15 \\
\bottomrule
\end{tabular*}
\end{table}

\myparagraph{Case study and 32B pilot.}
Fig.~\ref{fig:qualitative-recovery} shows an All-8B trajectory in which a
conflict between \texttt{image\_facts} and \texttt{reasoning\_steps} lets the
text-only Critic trigger a targeted recheck that repairs the observation of
option A. We present it as an illustration of the mechanism rather than a
typical outcome, as most Board updates maintain or extend the evidence instead of correcting an observation.

Complementing the 500-question ablations, our 50-question All-32B diagnostic
study in the supplement isolates sharing, history compression, and tool
interfaces. Sharing the Board with the Critic improves accuracy from 60.00\%
to 64.00\% and lowers mean latency by 24.3\%.
Selective Board updates also outperform an exhaustive enumeration prompt,
which reaches 58.00\%, while disabling the Board entirely drops accuracy to
42.00\%, a 22.00\pp{} loss that is the largest single effect in this study.
Enabling Board-preserving history compression raises accuracy from 56.00\%
to 64.00\% and cuts mean time by 31.0\%, showing
that pruning stale tool transcripts prevents context dilution. Among tool
interfaces, region-focused Crop reaches 66.00\% and outperforms
whole-image Query alone at 58.00\%, while providing both reaches
64.00\% and runs 31.0\% faster than Query alone.

\subsection{When Does Text-Only Critique Help?}
\label{subsec:critic}
Tab.~\ref{tab:critic} compares configurations with and without Critic
feedback while retaining a Board and visual tools. The no-Critic
configuration reduces accuracy from 48.80\% to 36.80\% for All-4B
and from 58.75\% to 50.80\% for All-8B, differences of 12.00 and 7.95\pp.
Flash exhibits the opposite pattern, reaching \textbf{69.80\%} without
the Critic compared with 68.47\% with it, while using 59.8\% fewer tokens
and 57.6\% less time. In this profile, removing the Critic also disables
post-submission visual checks.
The Critic benefit thus shifts from +12.00\pp{} on All-4B and
+7.95\pp{} on All-8B to $-1.33$\pp{} on Flash, reflecting backbone reasoning
capacity. Smaller VLMs often overlook premise--answer conflicts that text
audit catches, whereas stronger Solvers stay consistent over the structured
Board and skip extra review rounds that add tokens without accuracy gains,
so the Critic is most useful for smaller Solvers such as All-4B and All-8B.

\subsection{Can the Board Strategy Be Learned? Fine-Tuning and Transfer}
\label{subsec:sft-results}
Fig.~\ref{fig:sft-transfer} compares the base and SFT Qwen3-VL-4B Solvers under Direct
QA and critic-free tool use.
On Euclid30K, before SFT, multi-turn tool use without a Critic lowers
accuracy relative to single-pass Direct QA, from 39.40\% to 31.20\% with
free-form notes and to 36.80\% with the default Board, as the base 4B model struggles
with multi-turn state updates. After 200 SFT steps, Direct QA
rises by 3.40\pp{} from 39.40\% to 42.80\%, whereas free-form and default-Board
tool use gain 24.00 and 19.00\pp{} to reach 55.20\% and 55.80\%, exceeding
SFT Direct QA by 12.40 and 13.00\pp.

On ChartQA-Bar,
SFT improves Direct QA by 12.83\pp, free-form tool use by
6.54\pp, and default-Board tool use by 4.36\pp.
The two critic-free tool-use configurations reach 83.78\% and
82.81\%, respectively, surpassing the 80.7\% of RL-trained
VTool-R1-7B~\citep{wu2026vtool} on the same 826-item split even though our
Solver never sees a chart during training.
This cross-backbone comparison is only indicative.

\myparagraph{SFT transfers the Board strategy to free-form notes.}
Trained solely on structured Board rollouts, SFT narrows the
default-over-free-form gap from 5.60 to 0.60\pp{} on Euclid30K and shifts
it from +1.21 to $-0.97$\pp{} on ChartQA-Bar. The Visual Board thus serves
both as a training-free scaffold and as supervision whose organization
strategy transfers to unconstrained notes.

\section{Conclusion}
\label{sec:conclusion}
We introduce a Visual Board-centered harness for multi-step visual reasoning that separates visual observations, uncertainties, and derived reasoning within a structured, capacity-bounded, and revisable workspace.
Rather than accumulating interaction history, the Board maintains an explicit reasoning state through controlled updates, with the original image remaining the authoritative source of visual evidence.

Across multiple model configurations and three benchmarks, our training-free harness consistently improves reasoning accuracy over standalone baselines.
Controlled ablations further demonstrate the advantages of structured Board organization over free-form notes and no-Board retries.
Supervised fine-tuning on harness trajectories also improves direct answering and tool-augmented inference, with positive transfer to chart reasoning.
Together, these findings highlight the importance of how visual information is organized, revised, and reused throughout multi-step reasoning, beyond retaining more context or adding tools and larger helper models.

\section{Limitations and Future Directions}
\label{sec:limitations}

Our harness adds inference cost through multi-round Solver, visual-tool, and Critic calls, and its text-only Critic can miss visually wrong but consistent premises.
Integrating lightweight System-1 verification, such as Jev-style decision models, could check visual claims directly against the image while reducing unnecessary reasoning and tool calls.
The Solver could also learn to edit its own context~\citep{shao2026clm} rather than follow a fixed Board schema.
Finally, our post-training uses only geometry trajectories; extending Board-aware learning to broader visual domains remains future work.

\FloatBarrier
\label{page:main-end}

\clearpage
{\small
\bibliographystyle{ieeenat_fullname}
\bibliography{references}
}
\end{document}